\documentclass[letterpaper]{article} 
\usepackage{aaai24}
\usepackage{times}  
\usepackage{helvet}  
\usepackage{courier}  
\usepackage[hyphens]{url}  
\usepackage{graphicx} 
\usepackage{natbib}  
\usepackage{caption} 
\usepackage{algorithm}
\usepackage{algorithmic}

\usepackage{amsmath, amsfonts, amsthm, stackrel, amssymb}
\usepackage{multirow}

\usepackage{newfloat}
\usepackage{listings}
\DeclareCaptionStyle{ruled}{labelfont=normalfont,labelsep=colon,strut=off} 
\floatstyle{ruled}
\newfloat{listing}{tb}{lst}{}
\floatname{listing}{Listing}
\title{Fine-Grained Annotation for Face Anti-Spoofing}
\author{
    Xu Chen\textsuperscript{\rm 1}, Yunde Jia\textsuperscript{\rm 2}, Yuwei Wu\textsuperscript{\rm 1,2}
}
\affiliations{
    \textsuperscript{\rm 1}Beijing Key Laboratory of Intelligent Information Technology,\\ School of Computer Science \& Technology, Beijing Institute of Technology, China\\
    \textsuperscript{\rm 2}Guangdong Laboratory of Machine Perception and Intelligent Computing, Shenzhen MSU-BIT University, China\\
    \{chenxu,jiayunde,wuyuwei\}@bit.edu.cn

}

\usepackage{bibentry}

\begin{document}

\maketitle

\begin{abstract}

Face anti-spoofing plays a critical role in safeguarding facial recognition systems against presentation attacks.
While existing deep learning methods show promising results, they still suffer from the lack of fine-grained annotations, which lead models to learn task-irrelevant or unfaithful features.
In this paper, we propose a fine-grained annotation method for face anti-spoofing.
Specifically, we first leverage the Segment Anything Model (SAM) to obtain pixel-wise segmentation masks by utilizing face landmarks as point prompts.
The face landmarks provide segmentation semantics, which segments the face into regions.
We then adopt these regions as masks and assemble them into three separate annotation maps: spoof, living, and background maps.
Finally, we combine three separate maps into a three-channel map as annotations for model training.
Furthermore, we introduce the Multi-Channel Region Exchange Augmentation (MCREA) to diversify training data and reduce overfitting.
Experimental results demonstrate that our method outperforms existing state-of-the-art approaches in both intra-dataset and cross-dataset evaluations.

\end{abstract}

\section{Introduction}

Face anti-spoofing (FAS) is crucial for the security of facial recognition systems.
Significant progress has been made in FAS, benefiting from deep learning techniques. Existing methods \cite{li2023learning,sun2023rethinking,liao2023domain,liu2022feature,yu2020searching,yu2020face,Liu2018,yang2019face} have delivered impressive results, identifying complex attack patterns and enhancing the overall security of face recognition systems. 

However, existing methods still suffer from the lack of fine-grained annotations, which leads models to learn task-irrelevant or unfaithful features.
We observed that models tend to overfit to task-irrelevant features, including lighting conditions, lighting conditions, variations in skin color, specific backgrounds, and objects outside the facial region. 
Another observation is that the model tends to learn unfaithful features that are not related to the human face. Examples include regions like hands, eyewear presence, and hair details.  
These task-irrelevant features and unfaithful features would misguide the model, leading to misclassification of live faces or attacks.
One of the main reasons for this problem is the absence of fine-grained annotations in the training datasets.

\begin{figure}[tb]
    \centering
    \includegraphics[width=.46\textwidth]{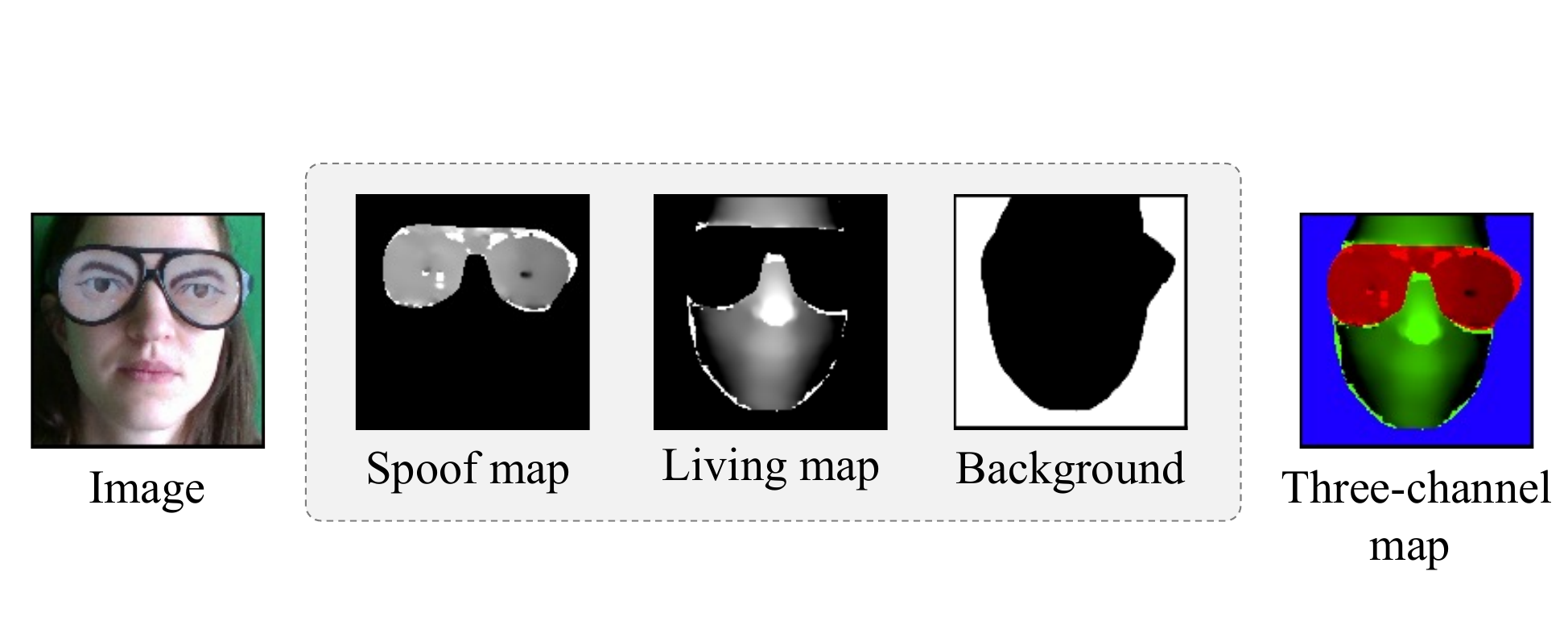}
    \centering
    \caption{
    Three-channel annotation map, representing attack, living face, and background regions. 
    }
    \label{Fig:fig1}
\end{figure}

Thanks to the Segment Anything Model (SAM) \cite{kirillov2023segment}, a model with a solid ability to generalize object segmentation can be used for fine-grained annotation.
The SAM has learned a general notion of what objects are, and this understanding enables zero-shot generalization to unfamiliar objects and images without requiring additional training.
Also, the promptable design enables flexible integration with other systems, which can take prompts to segment out precisely, providing an interactive annotation method.

In this paper, we propose a fine-grained annotation method for face anti-spoofing.
Firstly, we leverage the SAM to obtain pixel-wise segmentation masks by utilizing face landmarks as point prompts. 
The face landmarks provide   segmentation semantics, dividing a face into multiple regions.
Then, we adopt these regions as masks and assemble them into three separate annotation maps: spoof map, living map, and background map, as shown in Figure \ref{Fig:fig1}.
The fine-grained pixel-level annotation should include: (1) Both attack and living faces should be labeled in an independent sample to enhance the model's robustness in learning; (2) Explicit annotations are needed for regions of the face to reduce unfaithful features and remove ambiguity; (3) Spatial structural information or auxiliary information to provide discriminative supervision signals.
Finally, we combine these annotations into a three-channel map as annotations for model training.
The three-channel map aims to ensure auxiliary supervision's trustworthiness by mitigating ambiguity and providing more explicit guidance for model training.
In addition, we introduce a Multi-Channel Region Exchange Augmentation (MCREA) to improve the diversity of training data and reduce overfitting. 

During training, the network learns to predict the three-channel map. In testing, the method leverage the predictions from different regions to distinguish between attacks and living faces.
We conduct extensive experiments on intra-dataset and cross-dataset, and results show that our method achieves state-of-the-art performance under most testing scenarios.
Moreover, we also describe the ablation study to further investigate the proposed method.

The main contributions are summarized as follows:
\begin{itemize}
\item we propose a novel method that leverages the Segment Anything Model (SAM) to achieve pixel-wise segmentation for fine-grained annotation.
\item We propose a Multi-Channel Region Exchange Augmentation method to improve the data diversity of FAS training.
\item We introduce a three-channel annotation map to achieve fine-grained annotation.
\end{itemize}

\section{Related Work}

\subsection{Traditional Face Anti-spoofing}

The early face anti-spoofing approaches adopt handcrafted descriptors LBP \cite{de2013lbp,Boulkenafet2015}, SIFT \cite{patel2016secure}, HOG \cite{Komulainen2013Context} and DoG \cite{Tan2010} for extracting effective spoofing patterns from various color spaces (RGB, HSV, and YCbCr).
Hybrid methods \cite{simonyan2014very,he2016deep,dosovitskiy2020image,8626161,8453011,rehman2020enhancing} extract handcrafted features and employ CNNs for semantic feature representation then with binary cross entropy loss.
Other approaches adopt 5-class CE Loss \cite{9946402}, word-wise CE loss \cite{mirzaalian2021explaining} for better performance.
Compared with handcrafted descriptors, we adopt fine-grained annotations provide better representation capacity and enhance the learning of intrinsic features.
Compared with binary labels, the fine-grained annotations contain low-level features to avoid the model only focusing on high-level pattern mining.

\subsection{Pixel-Wise Supervision for Face Anti-spoofing}

Recently, many approaches adopt auxiliary tasks as the prior knowledge to guide the feature learning toward more generalizable cues.
Many approaches based on popular pseudo-depth annotation \cite{Atoum2017a,Liu2018,kim2019basn,yu2020searching,wang2020deep,yu2020face,zhang2020face,yu2020fas,yu2021dual,9730902} that predict the real depth for living faces while zero maps for spoof ones.
Besides that adopt depth as an annotation, the FAS community improves performance by finding different auxiliary annotations, such as binary mask label \cite{george2019deep,liu2019deep,sun2020face}, pseudo reflection map \cite{yu2020face,kim2019basn}, 3D point cloud map \cite{li20203dpc}, and ternary map \cite{sun2020face}.
However, 3D attacks make the auxiliary annotation not always reliable.
To solve this problem, we extend single-channel to three-channel annotation to solve ambiguity between 2D and 3D attacks. 
Generative model \cite{li2023learning,liu2022spoof,qin2021meta,liu2020disentangling,Jourabloo2018} mine the visual spoof patterns existing in the spoof samples to provide intuitive interpretation.
The generate pixel-wise annotations as supervision might easily fall into the local optimum and overfit on unexpected interference \cite{yu2022deep}.
In this paper, we adopt fine-grained annotation to mitigate this problem.

\subsection{Data Augmentation for Face Anti-spoofing}
Data Augmentation to increase the diversity of data plays a vital role in FAS.
The data augmentation methods \cite{buslaev2020albumentations} were introduced to improve the diversity of data.
A mixing augmentation approach \cite{huang2022generalized} is proposed based on mixing domain-specific feature statistics from different instances.
The approach \cite{10098629} draws inspiration from the theoretical error bound of domain generalization to use negative data augmentation instead of real-world attack samples for training.
\cite{Li2018a} propose a data augmentation method based on video cubes to remarkably
increase the number of training data. 
\cite{9946402} introduced a style-transfer network to augment the training data,
where both positive and negative samples are augmented.
Compare with existing methods, we exchange regions with semantic in image to improve data diversity.
For target data exploration \cite{liu2022source}, a specified patch shuffle data augmentation to explore intrinsic spoofing features for unseen attack types.
\cite{yu2021dual} generates more diverse spoofing samples by exchanging patches of real and spoof images.
We adopt different regions for data Augmentation, which is different from patch enhancement in providing a more semantically informative way of attack samples.

\begin{figure*}[ht]
    \centering
    \includegraphics[width=1.\textwidth]{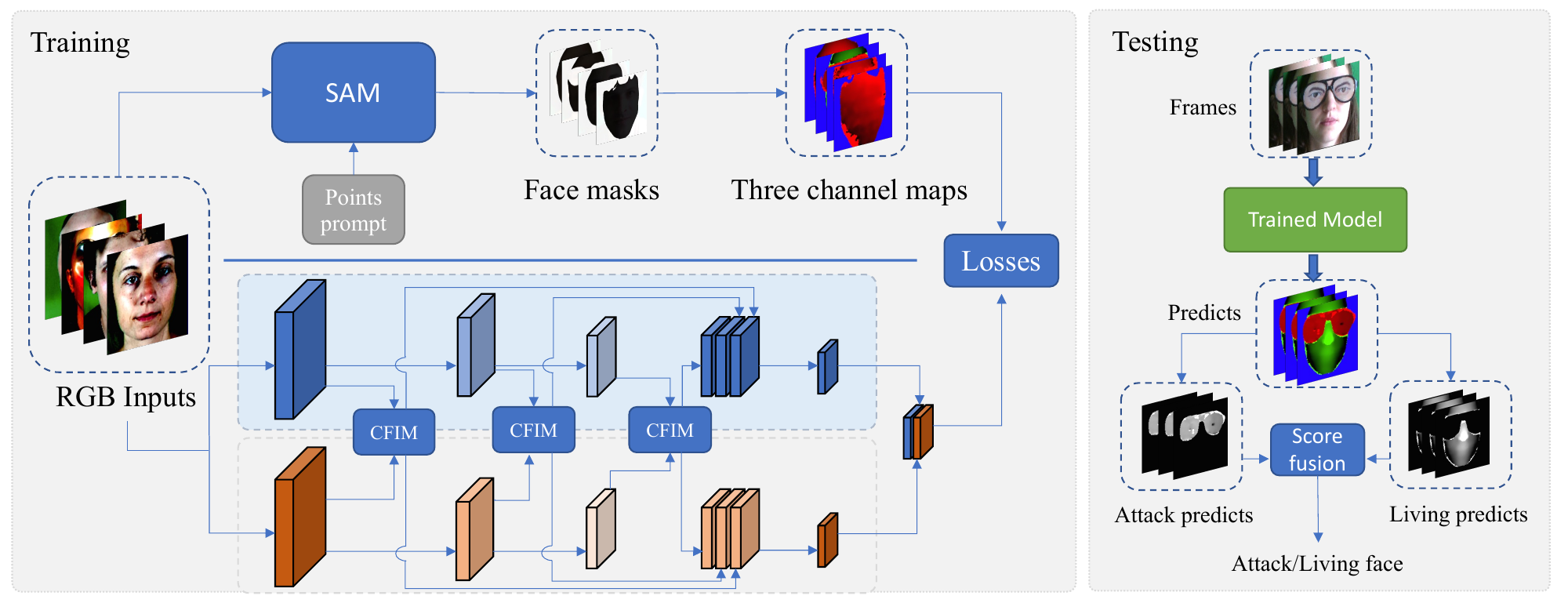}
    \centering
    \caption{An overview of the proposed method. 
    The input image is segmented by the SAM model with face landmarks as points prompt, resulting in three pixel-level maps representing attack, real, and background cues. 
    These maps are then used an annotations to train the model. During testing, the input image is passed through the model, which predicts the three-channel map. These predictions from different channels are fused to finally determine the output of the model. Cross Feature Interaction Modules (CFIM) in each level layer for mutual relation mining and local detailed representation enhancement.
    }
    \label{fig2}
\end{figure*}

\section{Method}

\subsection{Overview}
The input to our model is RGB images in a batch. 
Each image is first processed through a three-channel maps generation module. 
Firstly, the Segment Anything Model (SAM) is employed to produce a fine-grained segmentation mask. The SAM uses face landmarks as prompts to segment face into regions. Then, these separate regions corresponding to living face or an attack regions.
Specifically, we define the skin of face region as living, the hair, glasses, and these regions unrelated to the task as the background. Others are defined as attacks.
Finally, we adopt the labeled regions to generate the final three-channel map. 
The model is trained to predict the three-channel maps. 
In testing, the method leverage the predictions from different regions to distinguish between attacks and living faces.
These predictions from different channels are fused, and their scores are used to determine the model's final output. Figure \ref{fig2} illustrates the structure of the propose method.

\subsection{Three-Channel Map Construction}

The composition of attacks is diverse, which makes different attacks have their own unique patterns which cause the pixel-level ground-truth labels for spoof patterns are unavailable \cite{Jourabloo2018,li2023learning}.
However, presenting attack instruments (PAI) can be identified, for example, a fake eye, printed paper, a silicone mask, or special lighting.
Attacker use of different materials to reproduce the face images, these materials are common in life, such as paper, screens, masks, etc.
Therefore, if a model learned the general notion of what objects are could identify these PAI.
Thanks to the SAM model, that demonstrated ability with zero-shot generalization to unfamiliar objects and images without requiring additional training.

In our method, we adopt the SAM to accomplish the following:
1) to segment facial regions by using face landmarks as point prompts and 
2) to segment PAI from image. 
Firstly, different regions are used to reduce the noise of the annotation. We only adopt the skin of the face as the living face, while parts of other faces as the background which unrelated to the task. We then defined PAI regions based on the type of attack from the training data.
The fine-grained pixel-level annotation should include: 1) Both the attack and living faces should be labeled in an independent sample to enhance the model's robustness in learning; 2) Explicit annotations are needed for regions of the face to reduce unfaithful features and remove ambiguity; 3) Spatial structural information or auxiliary information to provide discriminative supervision signals.

\noindent\textbf{Face landmarks for Prompt Engineering}
The SAM model equipped a variety of prompt types, including boxes and points.
We adopt face landmarks as points prompt for obtaining robust segmentation masks. Let $\mathcal{L}_f = \{l_1, l_2, \ldots, l_n\}$ denote the set of face landmarks extracted from a face image $\mathbf{I}$, where each $l_i = (x_i, y_i)$ represents the position of a landmark. 

For SAM $\mathcal{S}$, the face landmarks $\mathcal{L}_f$ to generate points prompt $P$. This prompt $P$ contains the positions of the landmarks and it guides the model to focus on specific areas of the face that are typically involved in attacks and living faces. 
The exact format of the prompt is domain-specific and is generated through a mapping function $\mathcal{F}: \mathcal{L}_f \rightarrow P$.
$\mathcal{L}_f$ is defined as the different regions of the human face, such as the eyes, mouth, eyebrows, forehead, nose, ears, and hair.
Also, SAM generate multiple valid masks for ambiguous face landmark prompts. 
The SAM model $\mathcal{S}$ is then used to generate a pixel-wise segmentation mask $\mathbf{M}$:

$$
\mathbf{M} = \mathcal{S}(\mathbf{I}, P).
$$

From our observation, 2D attacks have different patterns with 3D attacks. 2D attack cues depend on the characteristics of the paper or screen. When using SAM for segmentation, screen boundaries or paper boundaries are easily be segmented, but features only in the face regions are not explicit, but exist in color patterns or special textures. This is a limitation of SAM for 2D attack.
In 3D attacks, PAI is explicit, which can be effectively segmented by SAM to achieve fine-grained annotations.

\begin{figure}[tb]
    \centering
    \includegraphics[width=.46\textwidth]{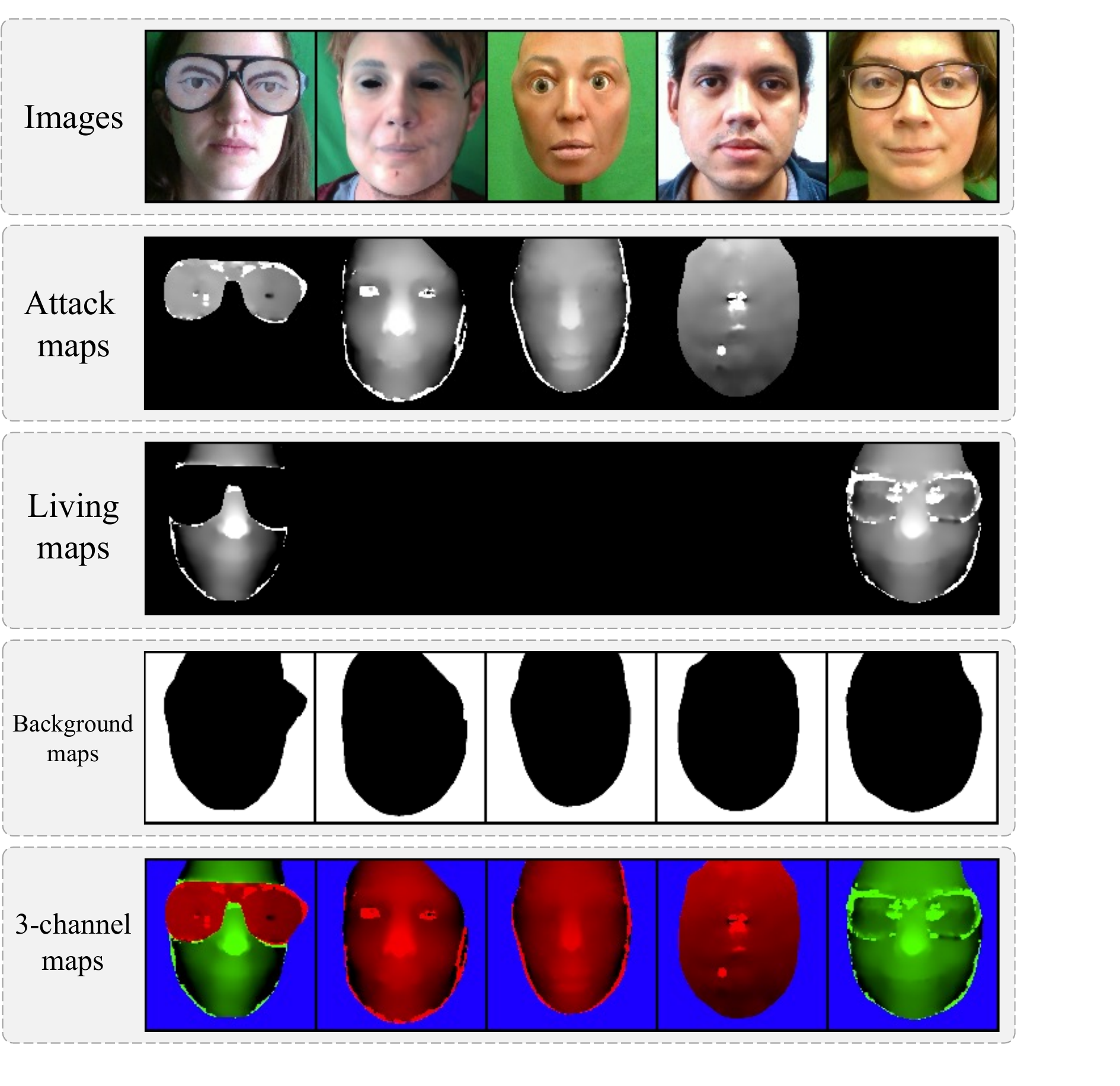}
    \centering
    \caption{
    An illustrative depiction of the three-channel map construction. Each channel within the map represents a distinct aspect of the input image: attack, living face, and background. The 'background' derived by the inversion of the union of the first two channels. 
    }
    \label{channel map}
\end{figure}

\noindent\textbf{Channel Pixel-wise Map}
The obtained segmentation regions are further processed for generating pixel-wise annotation.
As illustrated in Figure \ref{channel map}, The three-channel map represents the following categories - attack, living face, and background.
In a data-driven model, the living face is defined as the face skin, but this can cause the model to learn similar features and overfit into untrustworthy patterns. To solve this problem, the attack and the living face need to appear together in an annotation to disambiguate.
Existing methods only adopt single channel, which leads to bias in the model learning process.

Inspired by \cite{Jourabloo2018,liu2022spoof}, an input face image $\mathbf{I}$ can be formulated as $\mathbf{I}=\hat{\mathbf{I}}+T$, 
where $\hat{\mathbf{I}}$ refers to the live component and $T$ indicates the spoof.
Therefore, an attack can be defined as covering the face of a real person, and a reasonable setup is that the first channel is the attack and the second channel is the real face.

The image $\mathbf{I}$ is segmented into three masks $\mathbf{M}_{{attack}}$, $\mathbf{M}_{{real}}$, and $\mathbf{M}_{{bg}}$ for attack, living face, and background respectively. Each mask $\mathbf{M}_i$ where $i \in \{{attack}, {real}, {bg}\}$ is a binary matrix of the same size as the image, where a pixel at position $(u, v)$ is set to 1 if it belongs to the corresponding category, and 0 otherwise. 
Although we can directly use binary mask as annotation, we still need to introduce essential differences to provide more annotation content. Depth information is a widely used pixel-wise annotation, but it is not valid for 3D attacks, so when we expand to multi-channel annotation, depth can be used to provide meaningful supervisory signals.

For generating the first two channels, we then obtain the depth map $\mathbf{D}$ of the image. The depth map can be computed using any depth estimation method or depth channel in dataset. Let $\mathbf{D}(u, v)$ be the depth value at pixel location $(u, v)$ in the depth map. We normalize the depth values to the range [0, 1].
In our paper, the depth of the background is set to 0.
Mathematically, for each pixel $(u, v)$ in the image $\mathbf{I}$, the masks are defined as

\begin{equation}
\mathbf{M}_i(u, v) = \mathbf{D}(u, v).
\end{equation}

Finally, the three-channel pixel-wise annotation is

\begin{equation}
\mathbf{L} = [\mathbf{M}_{{attack}}, \mathbf{M}_{{real}}, \mathbf{M}_{{bg}}].
\end{equation}

\subsection{Model Architecture and Losses}

We adopt dual-cross central difference Network (DC-CDN) \cite{yu2020searching,yu2021dual} as backbone, which is established with Cross Feature Interaction Modules (CFIM) for mutual relation mining and local detailed representation enhancement.
The original DC-CDN network was used to predict a single channel of depth data, and we extended it to three channels. The predicted output has the same size as the input image.

Our method uses two types of loss functions, Mean Square Error (MSE) loss $L_{mse}$ and Contrastive Depth loss(CD) $L_{cd}$ \cite{wang2020deep}.
The MSE loss function ensures that the model produces a segmentation mask that closely matches the pixel-wise label, while the contrastive depth loss function ensures that the model correctly predicts the depth information present in the image.
The final loss function given by
\begin{equation}
L = \alpha L_{mse} + \beta L_{cd}.
\end{equation}

\subsection{Multi-Channel Region Exchange Augmentation}
Data Augmentation to increase the diversity of data plays a vital role in FAS. 
We proposed a data augmentation method to address the limited diversity and scale issues in FAS datasets. 
Thanks to the segmentation of SAM, regions with semantic exchange are conducive to enhancing data diversity.
The regions of faces contain semantic information, which is exchanged in a batch to obtain a new image. In order to keep the semantics of the generated images, face landmarks were used to align the regions to ensure semantic consistency.
The complete algorithm is summarized in Algorithm \ref{algoritm1}. 

\begin{figure}[htb]
    \centering
    \includegraphics[width=.5\textwidth]{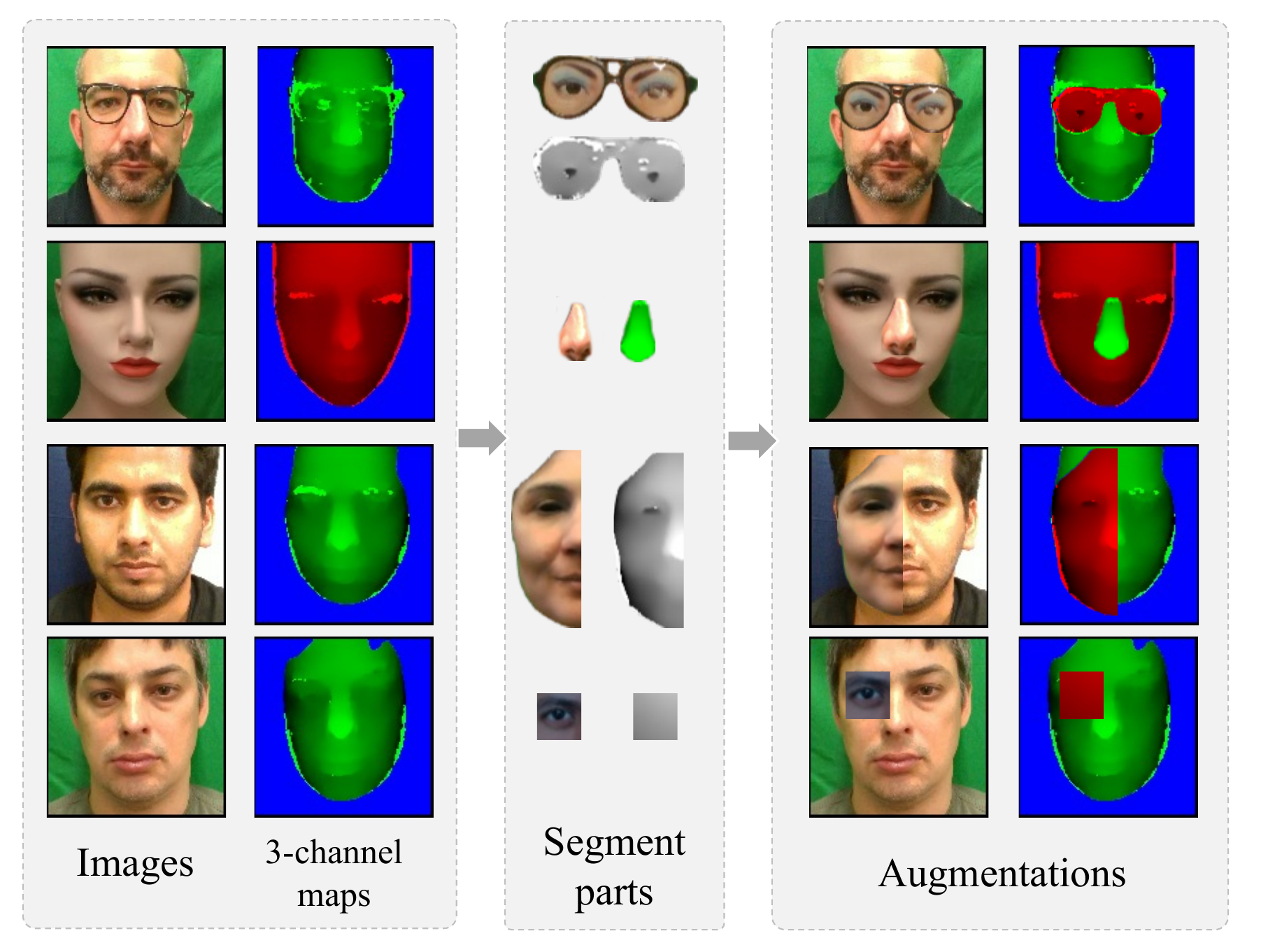}
    \centering
    \caption{
    Three data enhancement schemes, including integrated attack parts, overlay enhancement, and clipping exchange.
    }
    \label{cpe}
\end{figure}

There are three main advantages to this augmentation method include:
\textit{Enriched data distribution:} Our augmentation introduces face parts from various domains, enhancing the diversity of the data distribution.
\textit{Mimicking partial attacks:} The exchange of parts simulates a more diverse set of spoofing attacks, providing a more robust and realistic training set for the model.
\textit{Learning intricate spoofing patterns:} This method encourages the model to learn more detailed and intrinsic features for spoofing detection, improving the model's performance on complex spoofing attacks.
As shown in Figure \ref{cpe}. We shet three data enhancement schemes, including integrated attack regions, overlay enhancement, and clipping exchange.

\begin{algorithm}[tb]
\renewcommand{\algorithmicrequire}{\textbf{Input:}}
\renewcommand{\algorithmicensure}{\textbf{Output:}}
\caption{Multi-Channel Region Exchanges Augmentation }
\begin{algorithmic}[1]
\REQUIRE Images $I$ with batchsize N, Three-channel map labels $L$, augmented ratio $\gamma \in [0,1] $, step number $\rho$
\FOR {each $I_i$ and $L_i, i= 1,...$, $\lfloor\gamma * N \rfloor$}
\FOR {each step $\rho$}
\STATE Randomly select region $R_{RGB_i},L_i$ within $I_i$, with region landmarks $\mathcal{L}_{R_i}$
\STATE Randomly select a batch index $j,j \le N$
\STATE Randomly select region $R_{RGB_j},L_j$ within $I_j$, with region landmarks $\mathcal{L}_{R_j}$
\STATE Alignment $R_{RGB_i},L_i$ by $\mathcal{L}_{R_i}$ in $I_j$
\STATE Exchange part $I_i(R_{RGB_i}) = I_j(R_{RGB_j})$, $I_i(L_i) = I_j(L_j)$ and 
$I_i(\mathcal{L}_{R_i}) = I_j(\mathcal{L}_{R_j})$
\ENDFOR
\ENDFOR
\ENSURE augmented $I$ and $L$
\end{algorithmic}
\label{algoritm1}
\end{algorithm}

\subsection{Attack and Living Face Determination}
In this section, another essential topic is how to utilize the predictions from the three channels to make a final decision. 
Existing methods that use single-channel predictions can be effective for 2D attacks, but they become ambiguous when dealing with 3D or partial attacks, potentially increasing the false negative rate. Hence, we propose to incorporate predictions from both the attack map and the living face map.

Let $I_{attack}$ and $I_{real}$ represent the pixel intensities of the ``attack" and ``living " channels, respectively. Let $A_{face}$ denote the area of the whole face and $A_{key}$ represent the key regions, including the eyes, nose, and mouth. 
These areas are manually defined by face landmarks.
We then use $F_{attack}$ and $F_{real}$ 
representing the prediction intensities of attack and real cues, respectively. 
The $F_{attack}$ and $F_{real}$ computed by

\begin{equation}
F_{attack} = \frac{\sum_{(x,y) \in A_{face}} I_{attack}(x,y)}{\left|A_{face}\right|}
\end{equation}

\begin{equation}
F_{real} = \frac{\sum_{(x,y) \in A_{key}} I_{real}(x,y)}{\left|A_{key}\right|}
\end{equation}

\noindent where, $(x,y)$ denotes the pixel coordinates, while $\left|A_{face}\right|$ and $\left|A_{key}\right|$ represent the total number of pixels in the face area and key regions, respectively. Finally, we define the prediction function $F_{pred}$ as

\begin{equation}
F_{pred} =
\begin{cases} 
1 & \text{if } F_{attack} > F_{real} + \epsilon, \\
0 & \text{otherwise.}
\end{cases}
\end{equation}

\noindent where $\epsilon$ is a hyperparameter that balances the importance of the two channels. This strategy not only considers attack cues across the whole face area but also emphasizes the authenticity in the key facial regions, enabling effective handling of 2D, 3D, and partial attacks.

\begin{table*}[htbp]
  \centering
  \setlength{\tabcolsep}{1.6mm}
  \renewcommand\arraystretch{1.1}
    \begin{tabular}{c|c|c|c|c|c|c|c|c}
    \hline
    Method & Flexiblemask  & Replay & Fakehead  & Prints  & Glasses  & Papermask  & Rigidmask  & Mean±Std  \\
    \hline
    MC-ResNetDLAS & 33.3  & 38.5  & 49.6  & 3.8   & 41    & 47    & 20.6  & 33.4±14.9  \\
    Auxiliary & 13.2 & 12.5 & 47.3 & 32.2 & 23.7 & 13.9 & 40.4 & 26.2±14.1 \\
    MCCNN-OCCL-GMM  & 22.8  & 31.4  & 1.9   & 30    & 50    & 4.8   & 18.3  & 22.7±15.3  \\
    CDCN  & 12.1  & 8.69  & 42.7 & 30.1 & 11.7 & 11.9 & 30.4 & 21.1±13.2 \\
    RGBD-MH-BCE  & 33.7  & 1.0    & 3.1   & 1.7   & 37.6  & \textbf{1.0}     & 2.2   & 11.4±15.3  \\
    MC-PixBiS  & 49.7  & 3.7   & 0.7   & 0.1   & 16.0    & 0.2  & 3.4   & 10.5±16.7  \\
    CMFL-FAS & 12.4  & 1.0   & 2.5   & 0.7   & 33.5  & 1.8   & 1.7  & 7.6±11.2  \\
    MMDN & 19.0 & \textbf{0.5} & 2.3 & 0.7 & 10.0 & 0.7 & \textbf{0.6} & 4.8±6.6 \\
    FM-ViT & \textbf{3.6} & 0.7 & \textbf{0.0} & \textbf{0.0} & 12.0 & \textbf{0.4} & 0.7 & 2.5±4.4 \\
    
    \hline
    \textbf{Ours}  &  6.5  & 2.6   & 1.9  & 2.7  & \textbf{7.8}   & 5.3   & 3.2   &  4.3±2.0\\
    
    \hline
    \end{tabular}%
    \caption{
  Performance comparison about intra-dataset cross-type testing for unknown spoof attack detection. We use ACER (\%) under the unseen protocol of the WMCA dataset. 
  The results are obtained with a threshold computed for BPCER 1\% in the development set.
  }
  \label{tab:WMCA benchmark}%
\end{table*}%

\begin{table}[ht]
\centering
\setlength{\tabcolsep}{1.0mm}
\renewcommand\arraystretch{1.1}
    \begin{tabular}{c|c|c|c|c}
    \hline
          & Train & Test  & Train & Test \\
\cline{2-5}    Method & CASIA- & Replay & Replay & CASIA- \\
          & FASD  & Attack & Attack & FASD\\
    \hline
    RRRM & \multicolumn{2}{c|}{16.9} & \multicolumn{2}{c}{27.4} \\
    DR-UDA & \multicolumn{2}{c|}{15.6} & \multicolumn{2}{c}{34.2} \\
    CDCN & \multicolumn{2}{c|}{15.5} & \multicolumn{2}{c}{32.6} \\
    SFSNet & \multicolumn{2}{c|}{9.8} & \multicolumn{2}{c}{29.6} \\
    CDCN++ & \multicolumn{2}{c|}{6.5} & \multicolumn{2}{c}{29.8} \\
    DC-CDN & \multicolumn{2}{c|}{6.0} & \multicolumn{2}{c}{30.1} \\ 
    \hline
    \textbf{Ours}  & \multicolumn{2}{c|}{\textbf{5.2}} & \multicolumn{2}{c}{\textbf{23.2}} \\
    \hline
    \end{tabular}%
    \caption{Performance comparison about cross-dataset evaluation. We use HTER(\%) metric on the CASIA-FASD dataset and the Replay-Attack dataset.}
  \label{Tab: CASIA-FASD Replay-Attack}%
\end{table}%

\begin{table}[ht]
\setlength{\tabcolsep}{1.2mm}
\renewcommand\arraystretch{1.1}
  \centering
    \begin{tabular}{c|c|c|c|c}
    \hline 
    \multirow{2}{*}{Method} & \multicolumn{4}{c}{ Protocols }\\ 
    \cline{2-5} 
      & \uppercase\expandafter{\romannumeral1} & \uppercase\expandafter{\romannumeral2} & \uppercase\expandafter{\romannumeral3} & \uppercase\expandafter{\romannumeral4}\\ 
    \hline
    DeepPix  & 0.4 & 6.0 & 11.1±9.4 & 25.0±12.7 \\
    STASN & 1.9 &2.2 &2.8±1.6 &7.5±4.7 \\
    Auxiliary &  1.6 & 2.7 &2.9±1.5 &9.5±6.0 \\
    STDN & 1.1 & 1.9 & 2.8±3.3 & 3.8±4.2 \\
    CDCN++ & 0.2 & 1.3 & 1.8±0.7  & 5.0±2.9 \\
    DC-CDN  & 0.4  & 1.3 & 1.9±1.1 & 4.0±3.1 \\
    PatchNet & \textbf{0} & 1.2 & \textbf{1.2±1.3} & 2.9±3.0\\
    \hline
    \textbf{Ours}  & 0.1 &\textbf{1.1} & 1.4±1.21 & \textbf{2.8±2.6} \\
    \hline
    \end{tabular}%
    \caption{Performance comparison using ACER(\%) metric on Protocol \uppercase\expandafter{\romannumeral1} \& \uppercase\expandafter{\romannumeral2}\& \uppercase\expandafter{\romannumeral3}\& \uppercase\expandafter{\romannumeral4} of OULU-NPU datatset.}
  \label{tab: U-OULU-NPU pro1234}%
\end{table}%

\section{Experiments}

\begin{table*}[ht]
  \centering
  \setlength{\tabcolsep}{1.6mm}
  \renewcommand\arraystretch{1.1}
    \begin{tabular}{c|c|c|c|c|c|c|c|c}
    \hline
    Method & Flexiblemask  & Replay & Fakehead  & Prints  & Glasses  & Papermask  & Rigidmask  & Mean±Std  \\
    \hline
    \textbf{w/o} attack and background channel  &  12.5  & 7.4   & 28.6  & 26.7  & 9.2   & 11.3   & 22.7   & 15.5±9.6  \\
    \textbf{w/o} living and background channel  &  8.5  & 7.6   & 6.9  & 17.7  & 8.4   & 11.6   & 24.2   & 12.1±5.9 \\
    \textbf{w/o} background channel  &  7.2  & 2.6   & 3.2  & 2.6  & 7.8   & 7.3   & 11.4   & 6.0±3.1 \\
    \hline
    \textbf{w/} three channel binary mask  &  9.4  & 2.6   & 8.4  & 2.6  & 10.2   & 9.3   & 5.7   &  6.9±3.0\\
    \hline
    \textbf{Ours}  &  6.5  & 2.6   & 1.9  & 2.7  & 7.8   & 5.3   & 3.2   &  4.3±2.0 \\
    \hline
    \end{tabular}%
    \caption{
    The WMCA ablation experiment uses the LLO protocol. Different annotation settings are used, including different channels and data types.
  }
  \label{tab:WMCA Ablation}%
\end{table*}%

\subsection{Datasets and Metrics}
We evaluate our proposed method on four publicly available face anti-spoofing datasets, including WMCA \cite{george2019biometric}, and OULU-NPU \cite{Boulkenafet2017}, CASIA-FASD \cite{Zhang2012} and Replay-Attack \cite{Chingovska2012}. 
The WMCA dataset comprises eight distinct presentation attack in 2D and 3D scenarios, including glass, fake head, print, replay, rigid mask, flexible mask, paper mask, and wig attacks.
Other datasets contain 2D spoof attacks (print attacks and replay attacks) with different illumination conditions and background scenes. 
In WMCA dataset, we use the threshold is calculated at BPCER = 1\% on the validation set.
We report three kinds of metrics to evaluate methods, including Bonafide Presentation Classification Error Rate (BPCER), Attack Presentation Classification Error Rate (APCER), Average Classification Error Rate (ACER).

\subsection{Implementation Details} 
\noindent\textbf{Base model}
We adopted the DC-CDN (Dual-cross Central Difference Network) \cite{yu2021dual} as framework, with the modification of setting the final output channel to $3$. 

\noindent\textbf{Hyper-parameters}
The model training on 8x NVIDIA P40 GPU, with a batch size of $160$, and the Adam optimizer was utilized for 500 epochs. 
The learning rate is $0.002$ while the learning rate halves every $200$ epochs. 
Within the DC-DCN architecture, the parameter $\theta = 0.7$. 
Data augmentation techniques such as random horizontal flipping and random cropping were applied during the training process. 
The hyper-parameter of MCREA is $\gamma=0.5 $ and $ \rho=1$.

\noindent\textbf{Points prompt for SAM}
Initially, we employ a dense face alignment method \cite{guo2020towards} to obtain facial landmarks, which are then used to generate the living face mask by the SAM model for regions such as the eyes, mouth, eyebrows, forehead, nose, ears, and hair.
For attacks, we use different sets of facial landmarks corresponding to various attack positions. These different landmark sets are used to create distinct annotation maps for each specific attack scenario.

\noindent\textbf{Depth in three-channel map}
For the WMCA dataset, we directly use the depth channel data provided by the dataset. However, for other datasets, we generate pseudo depth maps using a dense face alignment method \cite{guo2020towards}.

\subsection{Evaluations}
\textbf{WMCA}
LOO Protocol: Result in Tab \ref{tab:WMCA benchmark}. The Leave-One-Out Protocol is specifically designed to assess the model's performance in detecting unseen attacks. 
For WMCA, the compared methods include MC-ResNetDLAS\cite{parkin2019recognizing}, Auxiliary\cite{Liu2018}, ResNet50\cite{he2016deep}, MCCNN-OCCL-GMM\cite{george2020learning}, CDCN\cite{yu2020searching}, RGBD-MH-BCE\cite{george2021cross}, MC-PixBiS\cite{george2019deep}, CMFL-FAS\cite{george2021cross}, FM-ViT\cite{liu2023fm}, MMDN\cite{li2023learning}.

Our method achieve comparable performance in results. In particular, in the glasses attack, we achieved the best performance.
Our model architecture only uses a simple convolutional network structure, and lacks the attention combination of global and local features, which may be the bottleneck to improve the performance of replay and print attacks.

\noindent\textbf{Cross-dataset testing}
We tested CASIA-FASD and Replay-Attack which contains 2D attacks mainly. Result in Figure \ref{Tab: CASIA-FASD Replay-Attack}.
In this setting, we compared methods include RRRM\cite{yao2020face}, DR-UDA\cite{wang2020unsupervised}, CDCN\cite{yu2020searching}, SFSNet\cite{pinto2020leveraging}, CDCN++\cite{yu2020searching} and DC-CDN \cite{yu2021dual}.
In the experiment, we found that SAM has limitations on the segmentation ability of low resolution. Although, we used MCREA to increase the diversity of training data, more research on the effectiveness of enhancements is still needed, which is worth exploring in future work.

\noindent\textbf{Intra-dataset testing}
We evaluate the performance of the method on the OULU-NPU dataset, result in Table \ref{tab: U-OULU-NPU pro1234}.
we compared methods include DeepPix \cite{george2019deep}, STASN \cite{yang2019face}, Auxiliary \cite{Liu2018}, STDN \cite{liu2020disentangling}, CDCN++ \cite{yu2020nasfas}, DC-CDN \cite{yu2021dual} and PatchNet \cite{wang2022patchnet}.
Our method achieves the best results on the protocol II\&IV.
In this experiment, we labeled skin as an attack to reduce occlusion due to hair. Depth information is also masked.
The fine-grained segmentation reduces non-face related information.
These results prove that our method is effective in 2D attack detection and can maintain the generalization performance to new subjects and environments.

\begin{table}[htbp]
\renewcommand\arraystretch{1.1}
  \centering
    \begin{tabular}{c|c|c}
    \hline
    Method & \textbf{w/o} MCREA   & \textbf{w/} MCREA  \\
    \hline
    Flexiblemask & 9.6  & 6.5    \\
    Replay &       6.3  & 2.6   \\
    Fakehead &     7.5  & 1.9    \\
    Prints &       8.7   & 2.7    \\
    Glasses &      11.2  & 7.8   \\
    Papermask &    10.4   & 5.3    \\
    Rigidmask &    7.5   & 3.2    \\
    \hline
    Mean±Std & 8.7±1.6 & 4.3±2.0  \\
    \hline
    \end{tabular}%
    \caption{Ablation study on the MCREA under the unseen protocol of WMCA dataset. 
  The results are obtained with the hyper-parameter of MCREA is $\gamma=0.5 $ and $ \rho=3$.}
  \label{tab: MCREA}%
\end{table}%

\subsection{Ablation Study}
\noindent\textbf{Different channel supervision}
The experiments were conducted to evaluate the effectiveness of three-channel map label, where we used different supervision signals for the model. We tested three scenarios: the attack channel, the living channel and without background channel. All these tested enabled MCREA.
These tests were performed using the LOO protocol on the WMCA dataset. Result in Table \ref{tab:WMCA Ablation}.
From the experimental results, we find that the intrinsic characteristics of the learning model between the attack channel and the living channel are different, which is manifested in the performance decline of fakehead attack when only living channel. In addition, background labeling helps the model learn the attack cues presented in the face while ignoring the shortcut features.

\noindent\textbf{Annotation type}
We compared the effectiveness of different auxiliary tasks by using depth as the supervision signal and binary mask as the supervision signal. Result in Table \ref{tab:WMCA Ablation}.
Binary mask lacks fine-grained spatial information, which causes its performance to degrade. But we found that this is related to the type of attack, and in 2D attacks, the performance is consistent with the use of deep data thanks to the supervision of the three-channel.
In addition, the WMCA dataset has multiple channels available for auxiliary information, and more experiments are presented in additional materials.

\noindent\textbf{Data augmentation} 
The performance improvement of MCREA is significant. 
We show the performance with and without MCREA in the Table \ref{tab: MCREA}.
The MCREA effectively improves the diversity of the data, which makes the enhanced data become more complex attack patterns, which reduces the overfitting of the model to untrustworthy patterns. 
Although MCREA improves the diversity of training data, region alignment still depends on the accuracy of existing algorithms, which may have an impact on the data enhancement process.
More on Data augmentation visualizations is shown in our additional material.

\begin{figure}[t]
    \centering
    \includegraphics[width=.43\textwidth]{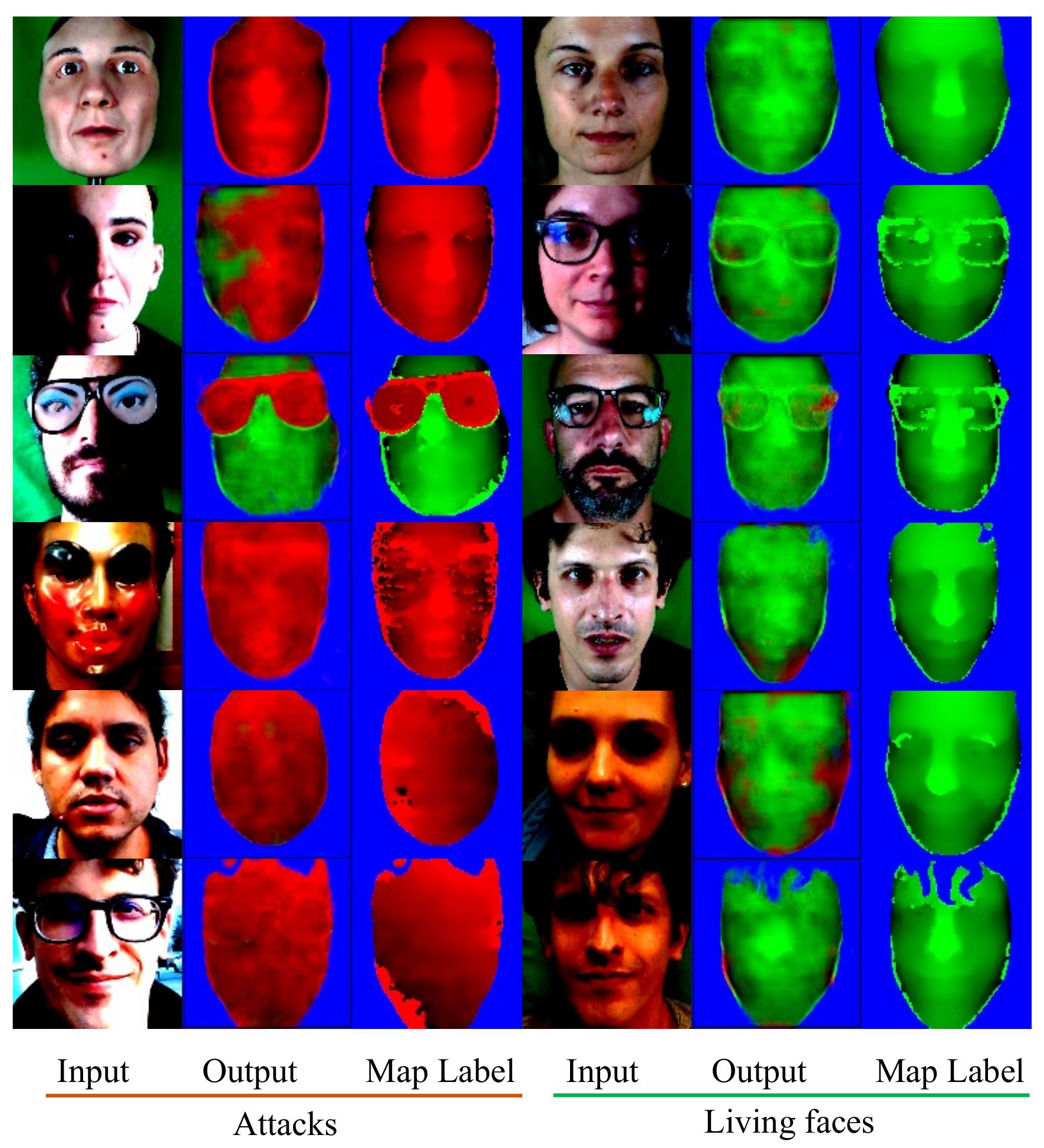}
    \centering
    \caption{The visualization of the three-channel prediction output by the model.}
    \label{Fig:output}
\end{figure}

\noindent\textbf{Visualization}
The model's predictions are shown in Figure \ref{Fig:output}.
We see that the shadow has a significant effect on the model, and the model shows wrong pixel prediction at different brightness boundaries.
This is because the model relies only on RGB image inputs that are still affected by light and shadow.

\section{Conclusion}

In this paper, we have presented a novel annotation method for face anti-spoofing.
We adopt the SAM to achieve pixel-wise segmentation for find-grained annotation.
We adopt SAM to segment faces into regions to enhance the diversity of the training data.
The experimental results show that our method achieve promising performance in both intra- and cross-dataset evaluations. 
We demonstrate the effectiveness of SAM for fine-grained annotation.
In addition, we verify that the performance of the model is improved by using SAM to segment the face to regions for data enhancement.
Our method there are still limitations, such as the ability of segmentation is still limited by SAM.

\bibliography{aaai24}

\end{document}